\pdfoutput=1

\documentclass[11pt]{article}

\usepackage[]{ACL2023}

\usepackage{times}
\usepackage{latexsym}
 \usepackage{graphicx, subcaption,ragged2e}
 \usepackage{multicol}
\usepackage{multirow}
\usepackage{booktabs}
\usepackage{makecell}

\usepackage[T1]{fontenc}

\usepackage[utf8]{inputenc}

\usepackage{microtype}
\usepackage[nameinlink,capitalize,noabbrev]{cleveref}

\usepackage{inconsolata}

\title{Text-To-KG Alignment:\\ Comparing Current Methods on Classification Tasks}

\author{Sondre Wold, Lilja Øvrelid, Erik Velldal \\  
  University of Oslo, Language Technology Group \\
  \texttt{\{sondrewo, liljao, erikve\}@ifi.uio.no} \\}

\begin{document}
\maketitle
\begin{abstract}
In contrast to large text corpora, knowledge graphs (KG) provide dense and structured representations of factual information. This makes them attractive for systems that supplement or ground the knowledge found in pre-trained language models with an external knowledge source. This has especially been the case for classification tasks, where recent work has focused on creating pipeline models that retrieve information from KGs like ConceptNet as additional context. Many of these models consist of multiple components, and although they differ in the number and nature of these parts, they all have in common that for some given text query, they attempt to identify and retrieve a relevant subgraph from the KG. Due to the noise and idiosyncrasies often found in KGs, it is not known how current methods compare to a scenario where the aligned subgraph is completely relevant to the query. In this work, we try to bridge this knowledge gap by reviewing current approaches to text-to-KG alignment and evaluating them on two datasets where manually created graphs are available, providing insights into the effectiveness of current methods. We release our code for reproducibility.\footnote{\url{https://github.com/SondreWold/graph_impact}}
\end{abstract}

\section{Introduction}
\begin{figure*}
    \centering
    \includegraphics[width=0.5\textwidth]{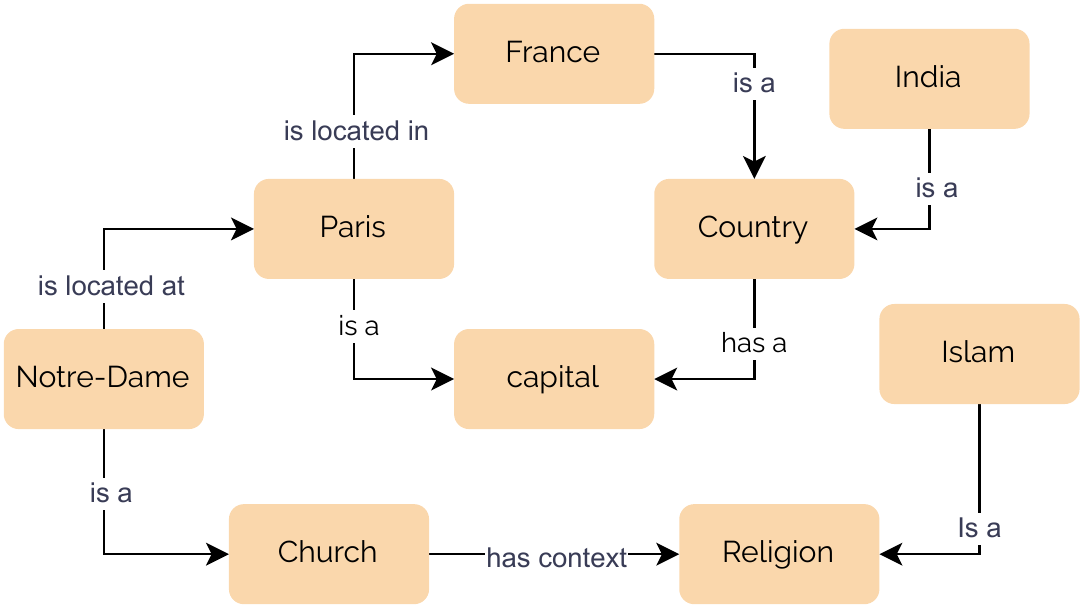}
    \caption{An example of a multi-relational knowledge graph.}
    \label{fig:paris}
\end{figure*}

There is a growing interest in systems that combine the implicit knowledge found in large pre-trained language models (PLMs) with external knowledge. The majority of these systems use knowledge graphs (KG) like ConceptNet \citep{speer2017conceptnet} or Freebase \cite{bollacker2008freebase} and either inject information from the graph directly into the PLM \citep{peters-etal-2019-knowledge, chang-etal-2020-incorporating, wang-etal-2020-connecting, lauscher-etal-2020-common, kaur-etal-2022-lm} or perform some type of joint reasoning between the PLM and the graph, for example by using a graph neural network on the graph and later intertwining the produced representations \citep{sun-etal-2022-jointlk, yasunaga-etal-2021-qa, greaselm, yasunaga2022deep}. Beyond their competitive performance, these knowledge-enhanced systems are often upheld as more interpretable, as their reliance on structured information can be reverse-engineered in order to explain predictions or used to create reasoning paths.

One of the central components in these systems is the identification of the most relevant part of a KG for each natural language query. Given that most KGs are noisy and contain idiosyncratic phrasings, which leads to graph sparsity \citep{sun-etal-2022-jointlk, jung-etal-2022-learning}, it is non-trivial to align entities from text with nodes in the graph. Despite this, existing work often uses relatively simple methods and does not isolate and evaluate the effect of this component on the overall classification pipeline. Furthermore, due to the lack of datasets that contain manually selected relevant graphs, it is not known how well current methods perform relative to a potential upper bound where the graph provides a structured explanation as to why the sample under classification belongs to a class. Given that this problem applies to a range of typical NLP tasks, and subsequently can be found under a range of different names, such as grounding, etc., there is much to be gained from reviewing current approaches and assessing their effect in isolation.  

In this paper, we address these issues by providing an overview of text-to-KG alignment methods. We also evaluate a sample of the current main approaches to text-to-KG alignment on two downstream NLP tasks, comparing them to manually created graphs that we use for estimating a potential upper bound. For evaluation, we use the tasks of binary stance prediction \citep{saha-etal-2021-explagraphs}, transformed from a graph generation problem in order to get gold reference alignments, and a subset of the Choice of Plausible Alternatives (COPA) \citep{roemmele2011choice} that contain additional explanation graphs \citep{brassard-etal-2022-copa}. As the focus of this work is not how to best combine structured data with PLMs, but rather to report on how current text-to-KG alignment methods compare to manually created graphs, we use a rather simple integration technique to combine the graphs with a pre-trained language model. Through this work, we hope to motivate more research into methods that align unstructured and structured data sources for a range of tasks within NLP, not only for QA.

\begin{figure*}
    \centering
    \includegraphics[width=0.7\textwidth]{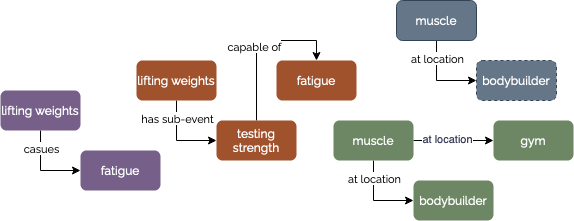}
    \caption{An example of the different graph construction approaches for COPA-SSE \citep{brassard-etal-2022-copa}. Here, the premise and answer options are: \textit{P: The bodybuilder lifted weights}; \textit{A1: The gym closed}; \textit{A2: Her muscles became fatigued}, from left to right: Purple: Gold annotation, Brown: Approach 3, Green: Approach 2, and Blue: Approach 1.}
    \label{fig:levels}
\end{figure*}

\section{Background}

Combining text with structured knowledge is a long-standing challenge in NLP. While earlier work focused more on the text-to-KG alignment itself, using rule-based systems and templates, recent work often approaches the problem as a part of a system intended for other NLP tasks than the alignment itself, such as question answering \citep{yasunaga-etal-2021-qa}, language modelling \citep{kaur-etal-2022-lm} and text summarization \citep{feng2021incorporating}.

As a consequence, approaches to what is essentially the same problem, namely to align some relevant subspace of a large KG with a piece of text, can be found under a range of terms, such as: \textit{retrieval} \citep{feng2021incorporating, kaur-etal-2022-lm, sun-etal-2022-jointlk, wang-etal-2020-connecting}, \textit{extraction} \citep{huang-etal-2021-improving, feng-etal-2020-scalable}, \textit{KG-to-text-alignment} \citep{agarwal-etal-2021-knowledge}, \textit{linking} \citep{gao-etal-2022-comfact, becker-etal-2021-coco}, \textit{grounding} \citep{shu-etal-2022-tiara, lin-etal-2019-kagnet}, and \textit{mapping} \citep{yu-etal-2022-KG}. Although it is natural to use multiple of these terms to describe a specific technique, we argue that it would be beneficial to refer to the task itself under a common name and propose the term \textit{text-to-KG alignment}. The following sections formalise the task and discuss current approaches found in the literature.

\subsection{Task definition}
The task of text-to-KG alignment involves two input elements: a piece of natural text and a KG. The KG is often a multi-relational graph, $G=(V, E)$, where $V$ is a set of entity nodes and $E$ is the set of edges connecting the nodes in $V$. The task is to align the text with a subset of the KG that is relevant to the text. What defines relevance is dependent on the specific use case. For example, given the question \textit{Where is the most famous church in France located?} and the KG found in \cref{fig:paris}, a well-executed text-to-KG alignment could, for example, link the spans \textit{church} and \textit{France} from the text to their corresponding entity nodes in the KG and return a subgraph that contains the minimal amount of nodes and edges required in order to guide any downstream system towards the correct behaviour. 

\subsection{Current approaches}
Although the possibilities are many, most current approaches to text-to-KG alignment base themselves on some form of lexical overlap. As noted in \citet{aglionby-teufel-2022-identifying, becker-etal-2021-coco, sun-etal-2022-jointlk}, the idiosyncratic phrasings often found in KGs make this problematic. One specific implementation based on lexical overlap is the one found in \citet{lin-etal-2019-kagnet}, which has been later reused in a series of other works on QA without any major modifications \citep{feng-etal-2020-scalable, yasunaga-etal-2021-qa, greaselm, yasunaga2022deep, sun-etal-2022-jointlk}.

In the approach of \citet{lin-etal-2019-kagnet}, a schema graph is constructed from each question-answer pair. The first step involves recognising concepts mentioned in the text that exists in the KG. Although they note that exact n-gram matches are not ideal, due to idiosyncratic phrasings and sparsity, they do little to improve on this naive approach besides lemmatisation and filtering of stop words, leaving it for future work. The enhanced n-gram matching produces two sets of entities, one from the question and one from the answer, $V_q$ and $V_a$. The graph itself is then constructed by adding the $k$-hop paths between the nodes in these two sets, with k often being $2$ or $3$. This returns a graph that contains a lot of noise in terms of irrelevant nodes found in the $k$-hop neighbourhoods of $V_q$ and $V_a$ and motivates some form of pruning applied to $G_{sub}$ before it is used together with the PLM, such as node relevance scoring \citep{yasunaga-etal-2021-qa}, dynamic pruning via LM-to-KG attention \citep{kaur-etal-2022-lm}, and ranking using sentence representations of the question and answer pair and a linearized version of $G_{sub}$ \citep{kaur-etal-2022-lm}.

Another approach based on lexical matching is from \citet{becker-etal-2021-coco}, which is specifically developed for ConceptNet. Candidate phrases are first extracted from the text using a constituency parser, limited to noun, verb and adjective phrases. These are then lemmatized and filtered for articles, pronouns, conjunctions, interjections and punctuation. The same process is also applied to all the nodes in ConceptNet. This makes it possible to match the two modalities better, as both are normalised using the same pre-processing pipeline. Results on two QA dataset show that the proposed method is able to align more meaningful concepts and that the ratio between informative and uninformative concepts are superior to simple string matching. For the language modelling task, \citet{kaur-etal-2022-lm} uses a much simpler technique where a Named Entity Recognition model identifies named entity mentions in text and selects entities with the maximum overlap in the KG. 

For the tasks of text summarisation and story ending generation, \citet{feng2021incorporating} and \citet{Guan_Wang_Huang_2019} use RNN-based architectures that read a text sequence word by word, and at each time step the current word is aligned to a triple from ConceptNet (We assume by lexical overlap). Each triple, and also its neighbours in the KG, is encoded using word embeddings and then combined with the context vector from the RNN using different attention style mechanisms. 

As an alternative to these types of approaches based on some form of lexical matching for the alignment, \citet{aglionby-teufel-2022-identifying} experimented with embedding each entity in the KG using a PLM, and then for each question answer pair find the most similar concepts using euclidean distance. They conclude that this leads to graphs that are more specific to the question-answer pair, and that this helps performance in some cases. \citet{wang-etal-2020-connecting} also experimented with using a PLM to generate the graphs instead of aligning them, relying on KGs such as ConceptNet as a fine-tuning dataset for the PLM instead of as a direct source during alignment. In a QA setting, the model is trained to connect entities from question-answer pairs with a multi-hop path. The generated paths can then be later used for knowledge-enhanced systems. This has the benefit of being able to use all the knowledge acquired during the PLMs pre-training, which might result in concepts that are not present in KGs.

\section{KG and Datasets}
This section explains the data used in our own experiments. 

\paragraph{ConceptNet}
As our knowledge graph, we use \textit{ConceptNet} \citep{speer2017conceptnet} --- a general-domain KG that contains $799,273$ nodes and $2,487,810$ edges. The graph is structured as a collection of triples, each containing a head and tail entity connected via a relation from a pre-defined set of types. 

\begin{figure}
    \centering
    \includegraphics[width=0.3\textwidth]{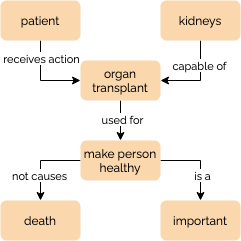}
    \caption{An example graph from ExplaGraphs \citep{saha-etal-2021-explagraphs} generated by a PLM for the belief argument pair: \textit{Organ transplant is important}; \textit{A patient with failed kidneys might not die if he gets organ donation}.}
    \label{fig:expla_example}
\end{figure}

\paragraph{ExplaGraphs}
ExplaGraphs \citep{saha-etal-2021-explagraphs} is originally a graph generation task for binary stance prediction. Given a belief and argument pair \textit{(b,a)}, models should both classify whether the argument counters or supports the belief and construct a structured explanation as to why this is the correct label. An example of this can be seen in \cref{fig:expla_example}.

The original dataset provides a train ($n=2367$) and validation ($n=397$) split, as well as a test set that is kept private for evaluation on a leaderboard. The node labels have been written by humans using free-form text, but the edge labels are limited to the set of relation types used in ConceptNet. We concatenate the train and validation split and partition the data into a new train, validation and test split with an 80--10--10 ratio. 

\paragraph{COPA-SSE}

\begin{figure}
    \centering
    \includegraphics[width=0.25\textwidth]{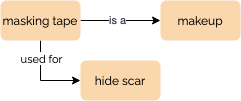}
    \caption{An example of a manually created graph from COPA-SSE \citep{brassard-etal-2022-copa} for the premise and options: \textit{P: The man felt ashamed of a scar on his face}; \textit{A1: He hid the scar with makeup}; \textit{A2: He explained the scar to strangers.}}
    \label{fig:copa_example}
\end{figure}

Introduced in \citet{brassard-etal-2022-copa}, COPA-SSE adds semi-structured explanations created by human annotators to 1500 samples from Balanced COPA \citep{kavumba-etal-2019-choosing} --- which is an extension to the original COPA dataset from \citet{roemmele2011choice}. In this task, given a scenario as a premise, models have to select the alternative that more plausibly stands in a causal relation with the premise. An example with a manually constructed explanation graph can be seen in \cref{fig:copa_example}. As with ExplaGraphs, COPA-SSE uses free-form text for the head and tail entities of the triples and limits the relation types to the ones found in ConceptNet. 

The dataset provides on average over six explanation graphs per sample. Five annotators have also rated the quality of each graph with respect to how well it captures the relationship between the premise and the correct answer choice. As we only need one graph per sample, we select the one with the highest average rating. As the official COPA-SSE set does not contain any training data, we keep the official development split as our training data and split the official test data by half for our in-house development and testing set. 

\section{Alignment approaches}
As mentioned, the general procedure for  grounding text to a graph is three-fold: we first have to identify entities mentioned in the text, then link them to entities in the graph, and lastly construct a graph object that is returned to the inference model as additional context to be used together with the original text. For QA the text aligned with the graph is typically a combination of the question and answer choices. As our two downstream tasks are not QA, and also different from each other, we have to rely on different pre-processing techniques than previous work. The following section presents the implementation of three different text-to-KG alignment approaches that we compare against manually created graphs. An illustration of the different approaches applied to the same text sample can be seen in \cref{fig:levels}.

\subsection{Approach 1: Basic String Matching}
Our first approach establishes a simple baseline based on naive string matching. For ExplaGraphs, we first word-tokenize the belief and argument on whitespace, and then for each word we check whether or not it is a concept in ConceptNet by exact lexical overlap. This gives us two sets of entities: $C_q$ and $C_a$. The graph is constructed by finding paths in ConceptNet between the concepts in $C_q$ and $C_a$. For COPA-SSE, we do the same but create $C_q$ from a concatenation of the premise and the first answer choice, and $C_a$ from a concatenation of the premise and the second answer choice. We use Dijkstra's algorithm to find the paths \cite{dijkstra1959note}.\footnote{Using the implementation from  \url{https://networkx.org}} The reason to use this rather simple approach, also pointed out by \citet{lin-etal-2019-kagnet} and \citet{aglionby-teufel-2022-identifying}, is that finding a minimal spanning graph that covers all the concepts from $C_q$ and $C_a$, which seems like a more obvious choice, would be to solve the NP-complete "Steiner tree problem" \citep{garey1977rectilinear}, and this would be too resource demanding given the size of ConceptNet. 

As many of the retrieved paths are irrelevant to the original text, it is common to implement some sort of pruning. We follow \citet{kaur-etal-2022-lm} and linearize the \texttt{subject-relation-object} triples to normal text and then embed them into the same vector space as the original context using the SentenceTransformer \cite{reimers-gurevych-2019-sentence}. We then calculate the cosine similarity between the linearized graphs and the original text context and select the one with the highest score.

\subsection{Approach 2: Enhanced String Matching}
Our second approach is based on the widely used method from \citet{lin-etal-2019-kagnet}, found in the works of \citet{feng-etal-2020-scalable, yasunaga-etal-2021-qa, greaselm, yasunaga2022deep, sun-etal-2022-jointlk}, but modified to our use case. We construct the set of entities $C_q$ and $C_a$ using n-gram matching enhanced with lemmatisation and filtering of stop words.\footnote{We use the implementation from \citet{yasunaga-etal-2021-qa} to construct $C_q$ and $C_a$} As in Approach 1, for ExplaGraphs, $C_q$ is constructed from the belief, and $C_a$ from the argument; for COPA-SSE, $C_q$ is based on a concatenation of the premise and the first answer choice, while $C_a$ is based on a concatenation of the premise and the second answer choice.

The graph is constructed by finding paths in ConceptNet from concepts in between $C_q$ and $C_a$ using the same method as in Approach 1. However, we limit the length of the paths to a variable $k$. In the aforementioned works, $k$ is set as to retrieve either two or three-hop paths, essentially finding the 2-hop or 3-hop neighbourhoods of the identified concepts. For our experiments, we set $k=3$.

As with Approach 1, many of the retrieved paths are irrelevant to the original text which warrants some sort of pruning strategy. In the aforementioned works, this is done by node relevance scoring. We follow Approach 1 and use sentence representations via linearization and cosine similarity in order to prune irrelevant paths from the graph.

\subsection{Approach 3: Path Generator}
Our third approach is based on a method where a generative LM is fine-tuned on the task of generating paths between concepts found in two sets. We use the implementation and already trained path generator (PG) from \citet{wang-etal-2020-connecting} for this purpose. This model is a GPT-2 model \citep{radford2019language} fine-tuned on generating paths between two nodes in ConceptNet. \footnote{See \citet{wang-etal-2020-connecting} for details on the fine-tuning procedure.}  One advantage of this method is that since GPT-2 already has unstructured knowledge encoded in its parameters from its original pre-training, it is able to generate paths between entities that might not exist in the original graph.

For both ExplaGraphs and COPA-SSE, we take the first and last entity identified by the entity linker from Approach 2 as the start and end points of the PG. As the model only returns one generated path, we do not perform any pruning. For the following example from COPA-SSE, \textit{P: The man felt ashamed of a scar on his face}; \textit{A1: He hid the scar with makeup}; \textit{A2: He explained the scar to strangers.}, the PG constructs the following path: \textit{masking tape used for hide scar, masking tape is a makeup}.

\subsubsection{Start and end entities}
We also experiment with the same setup, but with the first and last entity from the gold annotations as the start and end points for the PG. We do this to assess the importance of having nodes that are at least somewhat relevant to the original context as input to the PG. In our experiments, we refer to this sub-method as Approach 3-G.

\subsection{Integration technique} \label{sec:injection_technique}
As the focus of this work is not how to best combine structured data with PLMs, but rather to report on how current text-to-KG alignment methods compare to manually created graphs, we use a rather simple integration technique to combine the graphs with a pre-trained language model and use it uniformly for the different alignment approaches. We conjecture that the ranking of the different linking approaches with this technique would be similar to a more complex method for reasoning over the graph structures, for example using GNNs. By not relying on another deep learning model for the integration, we can better control the effect of the graph quality itself.

For each text and graph pair, we linearize the graph to text as in \citet{kaur-etal-2022-lm}. For example, the graph in \cref{fig:copa_example} is transformed to the string \textit{masking tape used for hide scar, masking tape is a makeup}. As linearization does not provide any natural way to capture the information provided by having directed edges, we transform all the graphs to undirected graphs before integrating them with the PLM \footnote{In practice, this is done by simply removing the underscore prepended to all reversed directions.}. For a different integration technique, such as GNNs, it would probably be reasonable to maintain information about the direction of edges.

For ExplaGraphs, which consists of belief and argument pairs, we feed the model with the following sequence: \textsc{belief [SEP] argument [SEP] graph [SEP]}, where \textsc{[SEP]} is a model-dependent separation token and the model classifies the sequence as either \textit{support} or \textit{counter}. 

For COPA-SSE, which has two options for each premise, we use the following format: \textsc{premise + graph [SEP] a1 [SEP]} and \textsc{premise + graph [SEP] a2 [SEP]}, where + just adds the linearized graph to the premise as a string and the model has to select the most likely sequence of the two.

\section{Graph quality}
The following section provides an analysis of the quality of the different approaches when used to align graphs for both ExplaGraphs and COPA-SSE.

 \cref{tab:stats_copa} and \cref{tab:stats_expla} show the average number of triples per sample identified or created by the different approaches for the two datasets, as well as how many triples we count as containing some form of error ('Broken triples' in the table). The criterion for marking a triple as broken includes missing head or tail entities inside the triple, having more than one edge between the head and tail, and returning nothing from ConceptNet. It is, of course, natural that not all samples contain an entity that can be found in ConceptNet, and consequently, we decided to not discard the broken triples but rather to include them to showcase the expected performance in a realistic setting.

 As can be seen from the tables, the approach based on the Path Generator (PG) from \citet{wang-etal-2020-connecting} (Approach 3) returns fewer triples than the other approaches for both ExplaGraphs and COPA-SSE. When using the entities from Approach 2 as the start and end points, denoted by the abbreviation Approach 3, the number of triples containing some form of alignment error is over twenty percent. When using the gold annotation as the start and end point of the PG, abbreviated Approach 3-G, this goes down a bit but is still considerably higher than the approaches based on lexical overlap. Approach 2 is able to identify some well-formatted triple in all of the cases for both tasks, while Approach 1 fails to retrieve anything for five percent of the samples in COPA-SSE and two percent for ExplaGraphs. 

In order to get some notion of semantic similarity between the different approaches and the original context they are meant to be a structural representation of, we calculate the cosine similarity between the context and a linearized (see \cref{sec:injection_technique} for details on this procedure) version of the graphs. The scores can be found in \cref{tab:similarities}. Unsurprisingly, the similarity increases with the complexity of the approach. The basic string matching technique of Approach 1 creates the least similar graphs, followed by the tad more sophisticated Approach 2, while the generative approaches are able to create a bit more similar graphs despite having a low number of average triples per graph. All of the approaches are still far from the manually created graphs --- which are also linearized using the same procedure as the others.

\begin{table}[!ht]
\small
  \centering
    \begin{tabular}{@{}lcc@{}}
    \toprule
    \textbf{Approach} & \textbf{Avg. number of triples} & \textbf{Broken triples} \\
    \midrule
    Approach 1 & $2.90$ & $0.05$ \\ 
    Approach 2 & $2.90$ & $0.00$ \\ 
    Approach 3 & $1.39$ & $0.20$ \\
    Approach 3-G & $1.64$ & $0.12$ \\
    \hline
    \noalign{\vskip 0.5mm}
    Gold & $2.12$ & $0.00$ \\ 
   \bottomrule
    \end{tabular}
    \caption{Statistics for the different approaches on the training set of COPA-SSE. The number of broken triples is reported as percentages.}
    \label{tab:stats_copa}
\end{table}

\begin{table}[!ht]
\small
  \centering
    \begin{tabular}{@{}lcc@{}}
    \toprule
    \textbf{Approach} & \textbf{Avg. number of triples} & \textbf{Broken triples} \\
    \midrule
    Approach 1 & $2.99$ & $0.02$ \\ 
    Approach 2 & $3.03$ & $0.00$ \\ 
    Approach 3 & $1.34$ & $0.21$ \\
    Approach 3-G & $1.58$ & $0.15$ \\
    \hline
    \noalign{\vskip 0.5mm}
    Gold & $4.23$ & $0.00$ \\ 
   \bottomrule
    \end{tabular}
    \caption{Statistics for the different approaches on the training set of ExplaGraphs. The number of broken triples is reported as percentages.}
    \label{tab:stats_expla}
\end{table}

\begin{table}[!ht]
\small
  \centering
    \begin{tabular}{@{}lcc@{}}
    \toprule
    \textbf{Approach} & \textbf{ExplaGraphs} & \textbf{COPA-SSE} \\
    \midrule
    Approach 1 & $0.39$ & $0.32$ \\ 
    Approach 2 & $0.45$ & $0.42$ \\ 
    Approach 3 & $0.48$ & $0.45$ \\
    Approach 3-G & $0.55$ & $0.46$ \\
    \hline
    \noalign{\vskip 0.5mm}
    Gold & $0.75$ & $0.57$ \\ 
   \bottomrule
    \end{tabular}
    \caption{The different graphs and their average cosine similarity with the original text.}
    \label{tab:similarities}
\end{table}

\section{Experiments}
We now present experiments where we compare the discussed approaches to text-to-KG alignment for ExplaGraphs and COPA-SSE. As our PLM, we use \textsc{bert} \citep{devlin-etal-2019-bert} for all experiments. We use the base version and conduct a hyperparameter grid search for both tasks. We do the same search both with and without any appended graphs as the former naturally makes it easier to overfit the data, especially since both ExplaGraphs and COPA-SSE are relatively small in size. The grid search settings can be found in \cref{sec_appendix_grid} and the final hyperparameters in \cref{sec_appendix_params}. We run all experiments over ten epochs with early stopping on validation loss with a patience value of five. 

As few-sample fine-tuning with \textsc{bert} is known to show instability \citep{zhang2021revisiting}, we run all experiments with ten random seeds and report the mean accuracy scores together with standard deviations. We use the same random seeds for both tasks; they can be found in \cref{sec_appendix_seeds}. 

We find that the experiments are highly susceptible to seed variation. Although we are able to match the performance of some previous work for the same PLM on some runs, this does not hold across seeds. Consequently, we also perform outlier detection and removal. Details on this procedure can be found in \cref{sec_appendix:outliers}.

\section{Results}
\begin{table}[t]
\small 
  \centering
    \begin{tabular}{@{}lcc@{}}
    \toprule
    \textbf{Approach} & \textbf{ExplaGraphs} & \textbf{COPA-SSE} \\
    \midrule
    No graph & $69.67^{\pm3.36}$ & $67.05^{\pm2.07}$ \\
    Approach 1 & $66.46^{\pm8.48}$ & $51.20^{\pm2.08}$ \\
    Approach 2 & $70.03^{\pm2.71}$ & $53.33^{\pm1.80}$ \\
    Approach 3 & $73.55^{\pm1.66}$ & $56.20^{\pm8.39}$ \\
    Approach 3-G & $70.57^{\pm3.27}$ & $85.86^{\pm0.75}$ \\
    \hline
    \noalign{\vskip 0.5mm}
    Gold & $80.28^{\pm2.31}$ & $96.60^{\pm0.28}$ \\ 
   \bottomrule
    \end{tabular}%
    \caption{Results of the different approaches on ExplaGraphs and COPA-SSE. Results are reported as average accuracy over ten runs together with standard deviations after outlier removal, if any.}
    \label{tab:results}
\end{table}

\begin{figure}
    \centering
    \includegraphics[width=0.3\textwidth]{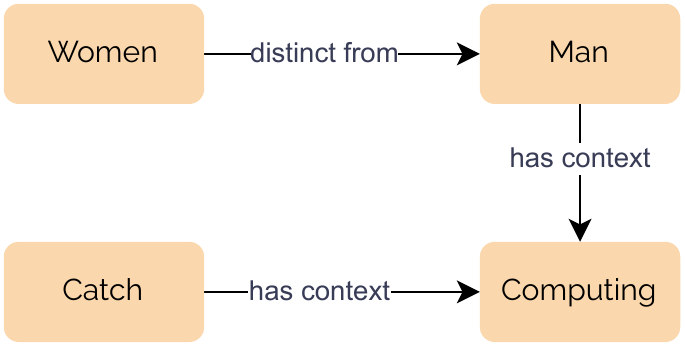}
    \caption{The graph aligned with ConceptNet for both the approaches based on lexical overlap. The original COPA-SSE context is \textit{Premise: The women met for coffee}  \textit{Alt 1: The cafe reopened in a new location};  \textit{Alt 2: They wanted to catch up with each other} }
    \label{fig:ret_linked}
\end{figure}

\cref{tab:results} shows the results on ExplaGraphs and COPA-SSE. For both datasets, we observe the following: Methods primarily based on lexical overlap provide no definitive improvement. The performance of Approach 1 (String matching) and Approach 2 (String matching with added lemmatisation and stop word filtering) is within the standard deviation of the experiments without any appended graph data, and might even impede the performance by making it harder to fit the data by introducing noise from the KG that is not relevant for the classification at hand. 

For Approach 3, based on a generative model, we see that it too provides little benefit for ExplaGraphs, but that when it has access to the gold annotation entities as the start and end point of the paths, it performs significantly better than having access to no graphs at all for COPA-SSE.

For both tasks, having access to manually created graphs improves performance significantly.
\begin{figure*}
    \centering
    \includegraphics[width=0.8\textwidth]{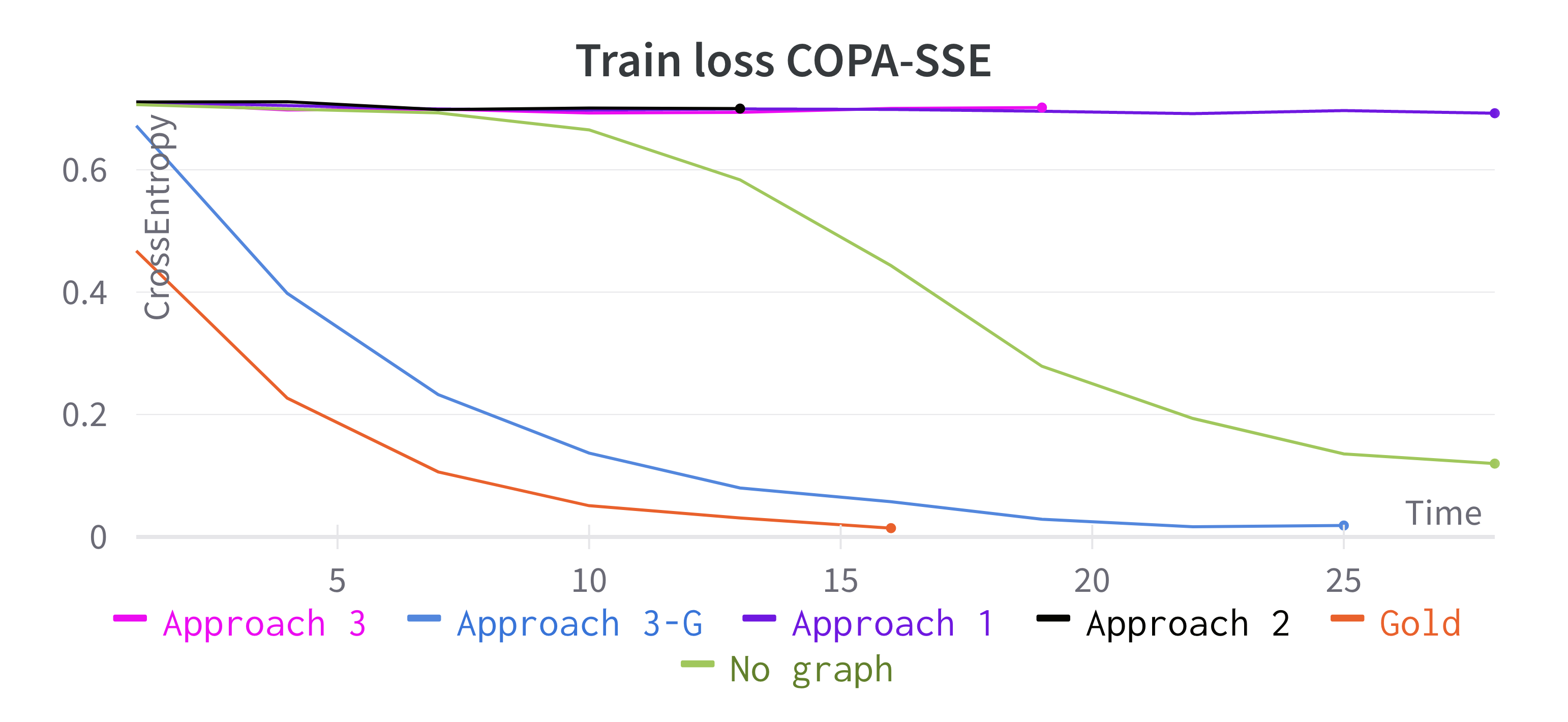}
    \caption{The train loss curves for the different approaches on COPA-SSE.}
    \label{fig:losses}
\end{figure*}

\section{Discussion}
The most striking result is perhaps the performance of Approach 3-G on COPA-SSE. We hypothesise that this can be explained by the fact that annotators probably used exact spans from both the premise and the correct alternative from the text in their graphs, and consequently, they provide a strong signal as to why there is a relation between the premise and the correct answer choice and not the wrong one. This is easily picked up by the model. For ExplaGraphs, which is a text classification problem, this is not the case: the appended graph might provide some inductive bias, but it does not provide a direct link to the correct choice, as the task is to assign a label to the whole sequence, not to choose the most probable sequence out of two options. This conclusion is further supported by the observation that appending the manually constructed graphs in their entirety has a much larger effect on COPA-SSE than ExplaGraphs.  

 Furthermore, for COPA-SSE, as pointed out in \cref{tab:stats_copa}, the average triple length for the generative approaches is rather low, so the majority of the aligned graphs from Approach 3-G are actually from the manually written text, not generated by the model itself. 

The key finding of our experiments is that having access to structured knowledge relevant to the sample at hand, here represented by the gold annotations, provides a significant increase in performance even with a simple injection technique and judging by today's standards, a small pre-trained language model. They also show that for datasets of low sample sizes, such as ExplaGraphs and COPA-SSE, the results are susceptible to noise. As the approaches based on lexical overlap are within the standard deviations of the experiments without any appended graphs, it is not possible to conclude that they add any useful information to the model. Based on \cref{fig:losses}, we think it is fair to conclude that these methods based on lexical overlap only provide a signal that has no relation to the correct label.
As to why the approaches based on lexical matching do not have any effect here but reportedly have an effect in previous work on QA, there is one major reason that has not been discussed so far: namely that both datasets require knowledge that is not represented in ConceptNet. As shown by \citet{bauer-bansal-2021-identify}, matching the task with the right KG is important. It is reasonable to question whether or not ConceptNet, which aims to represent commonsense and world knowledge, does indeed contain information useful for deciding whether or not an argument counters or supports a belief, in the case of ExplaGraphs, or if it can aid in the selection of the most likely follow-up scenario to a situation, in the case of COPA-SSE. In \cref{fig:ret_linked}, both the approaches based on lexical overlap (1 \& 2) align the same exact graph with the text context, and judging from the result, it is pretty clear that the aligned graph has little to offer in terms of guiding the model towards the most likely follow-up. 

\section{Conclusion}

In this work, we find that the process of identifying and retrieving the most relevant information in a knowledge graph is found under a range of different names in the literature and propose the term text-to-KG alignment.  
We systematise current approaches for text-to-KG alignment and evaluate a selection of them on two different tasks where manually created graphs are available, providing insights into how they compare to a scenario where the aligned graph is completely relevant to the text. Our experiments show that having access to such a graph could help performance significantly, and that current approaches based on lexical overlap are unsuccessful under our experimental setup, but that a generative approach using a PLM to generate a graph based on manually written text as start and end entities adds a significant increase in performance for multiple-choice type tasks, such as COPA-SSE. For the approaches based on lexical overlap, we hypothesise that the lack of performance increase can be attributed to the choice of knowledge graph, in our case ConceptNet, which might not contain any information useful for solving the two tasks.


\section*{Limitations}
While there is a lot of work on creating and making available large pre-trained language models for a range of languages, there is to our knowledge not that many knowledge graphs for other languages than English --- especially general knowledge ones, like ConceptNet. This is a major limitation, as it restricts research to one single language and the structured representation of knowledge found in the culture associated with that specific group of language users. Creating commonsense KGs from unstructured text is a costly process that requires financial resources for annotation as well as available corpora to extract the graph from.

\section*{Ethics Statement}
We do not foresee that combining knowledge graphs with pre-trained language models in the way done here, add to any of the existing ethical challenges associated with language models. However, this rests on the assumption that the knowledge graph does not contain any harmful information that might inject or amplify unwanted behaviour in the language model.

\bibliography{anthology,custom}
\bibliographystyle{acl_natbib}

\newpage
\appendix

\section{Appendix A}
\label{sec:appendix_B}

\subsection{SentenceTransformer}\label{sec_appendix_sentTrans}
We use the model with id \textsc{all-mpnet-base-v2} to prune the different paths and to calculate similarity.

\subsection{Grid search}\label{sec_appendix_grid}

Based on the following values, we do a grid search checking every possible combination. 

\begin{table}[!h]
    \centering
    \begin{tabular}{lc}
        \textbf{Hyperparameter} & \textbf{Value} \\ \hline
        \noalign{\vskip 0.5mm}
        lr & \makecell{$ 4* 10^{-5}$, $3 * 10^{-5}$ \\ $5 * 10^{-5}$, $6 * 10^{-6}$ \\ $4 * 10^{-6}$, $1 * 10^{-6}$}\\
        Weight decay & $0.01$ | $0.1$ \\
        Batch size & $4$ | $8$ | $16$ \\
        Dropout & $0.2$ | $0.3$ \\
    \end{tabular}
    \caption{The values used for the grid search}
    \label{tab:grid_search}
\end{table}

\subsection{Hyperparameters}\label{sec_appendix_params}

Based on the grid search, we select the following hyperparameters:

\begin{table}[!h]
    \centering
    \begin{tabular}{lcc}
        \textbf{Hyperparameter} & \textbf{With graphs} & \textbf{w/o graphs} \\ \hline
        \noalign{\vskip 0.5mm}
        Learning rate & $3 * 10^{-5}$ & $4 * 10^{-5}$ \\
        Dropout & $0.3$ & $0.3$ \\
        Weight decay & $0.01$ & $0.1$ \\
        Batch size & $16$ & $8$ \\
    \end{tabular}
    \caption{The hyperparameters used for ExplaGraphs}
    \label{tab:params_expla}
\end{table}

\begin{table}[!h]
    \centering
    \begin{tabular}{lcc}
        \textbf{Hyperparameter} & \textbf{With graphs} & \textbf{w/o graphs} \\ \hline
        \noalign{\vskip 0.5mm}
        Learning rate & $4 * 10^{-5}$ & $4 * 10^{-5}$ \\
        Dropout & $0.2$ & $0.3$ \\
        Weight decay & $0.01$ & $0.1$ \\
        Batch size & $8$ & $16$ \\
    \end{tabular}
    \caption{The hyperparameters used for COPA-SSE}
    \label{tab:params_copa}
\end{table}

\subsection{Seeds}\label{sec_appendix_seeds}

Seeds used for both tasks during fine-tuning: $[9, 119, 7230, 4180, 6050, 257, 981, 1088, 416, 88]$

\subsection{Outliers} \label{sec_appendix:outliers}

\begin{figure*}[!ht]
\captionsetup[subfigure]{justification=Centering}

\begin{subfigure}[t]{0.3\textwidth}
    \includegraphics[width=\textwidth]{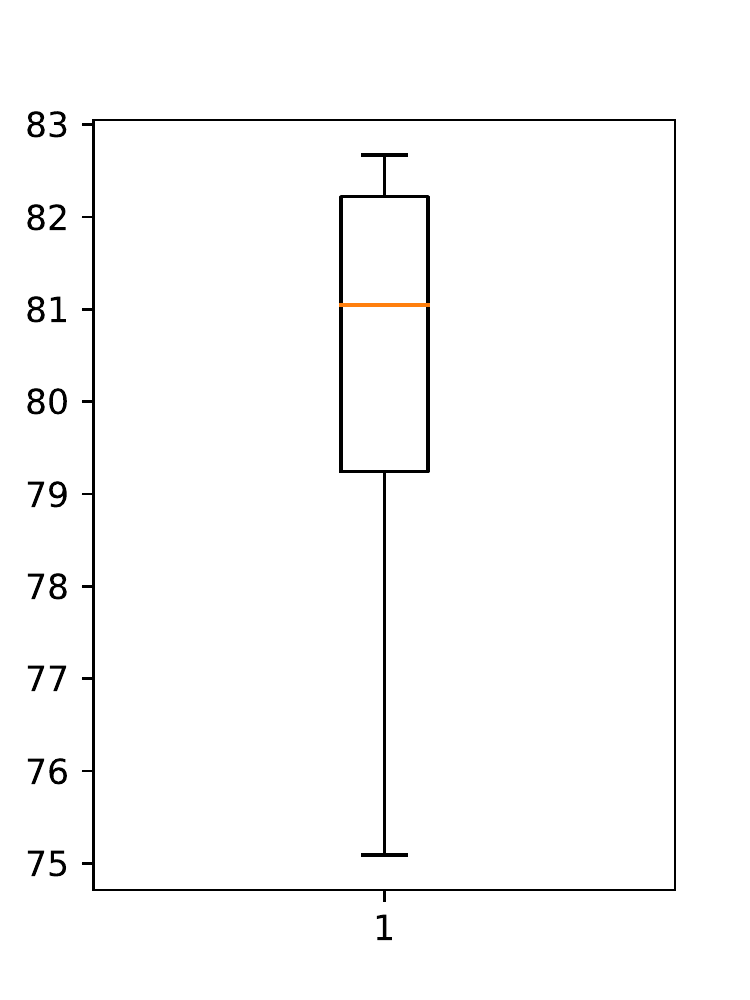}
    \caption{Manually created graphs}
\end{subfigure}\hspace{\fill} 
\begin{subfigure}[t]{0.3\textwidth}
    \includegraphics[width=\linewidth]{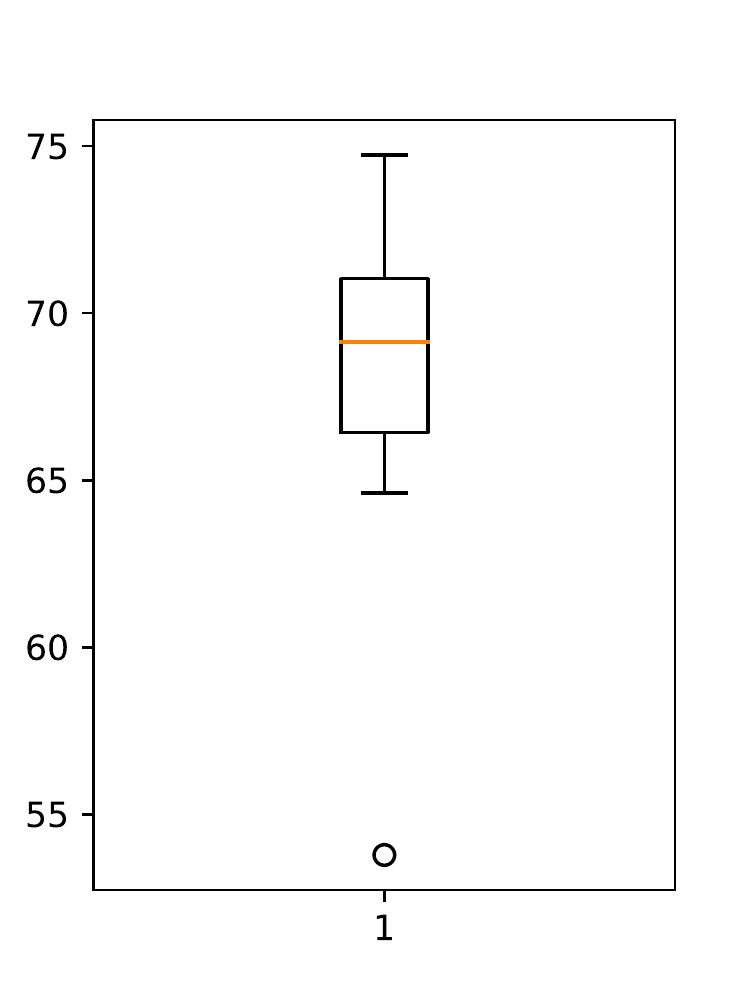}
    \caption{No graphs appended to original context}
\end{subfigure}\hspace{\fill}
\begin{subfigure}[t]{0.3\textwidth}
    \includegraphics[width=\linewidth]{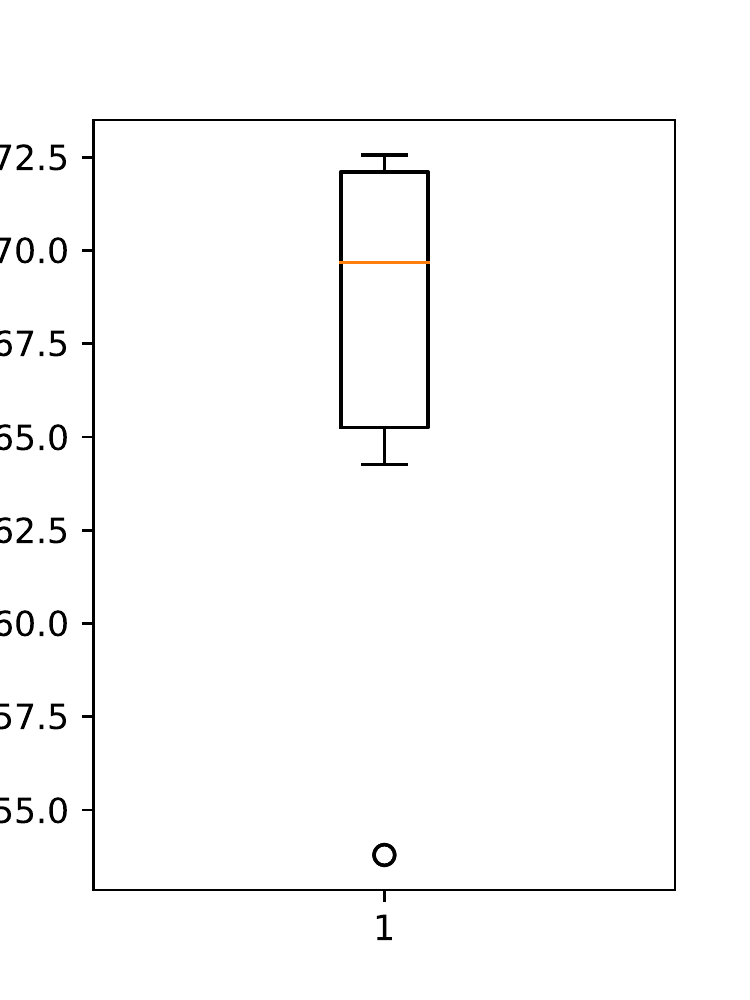}
    \caption{Approach 2}
\end{subfigure} 

\begin{subfigure}[t]{0.30\textwidth}
    \includegraphics[width=\linewidth]{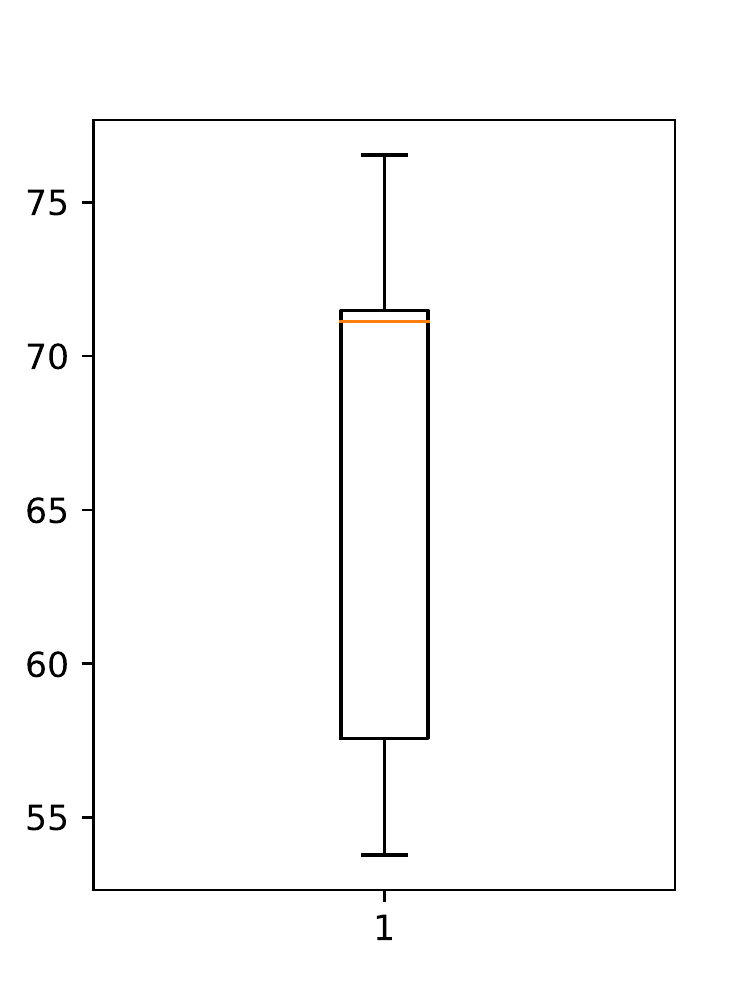}
    \caption{Approach 1}
\end{subfigure}\hspace{\fill}
\begin{subfigure}[t]{0.30\textwidth}
    \includegraphics[width=\linewidth]{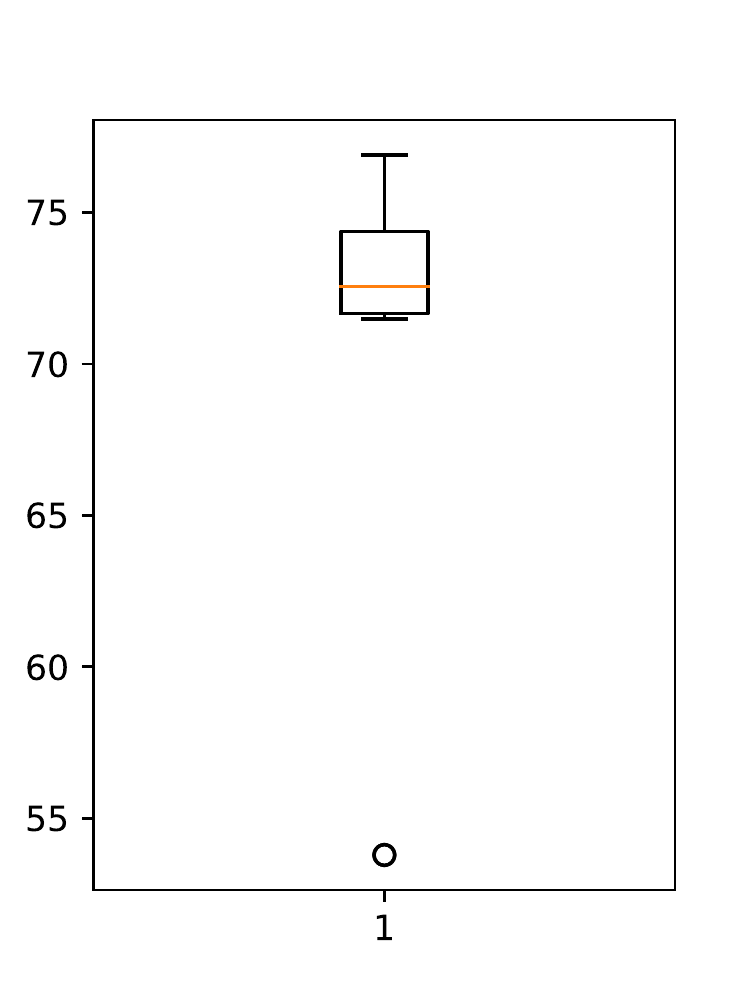}
    \caption{Approach 3}
\end{subfigure}\hspace{\fill} 
\begin{subfigure}[t]{0.30\textwidth}
    \includegraphics[width=\linewidth]{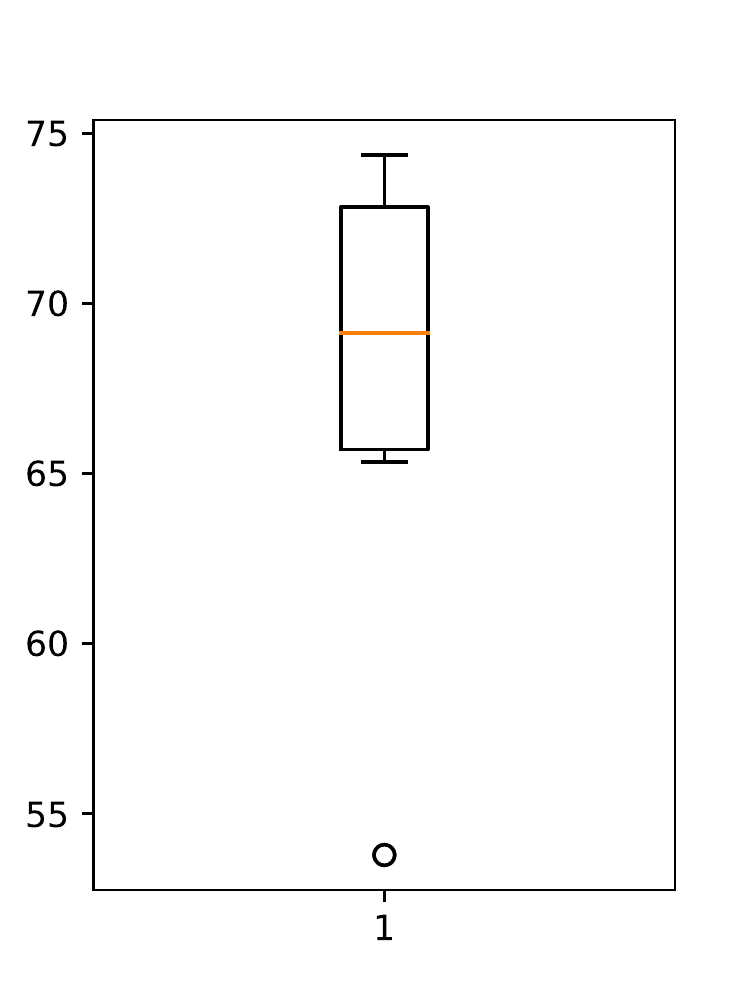}
    \caption{Approach 3-G}
\end{subfigure}

\caption{Outliers from the different runs for all graph configurations for ExplaGraphs. Circular dots mark outliers that were removed, if any.}

\end{figure*}

\begin{figure*}[!ht]
\captionsetup[subfigure]{justification=Centering}

\begin{subfigure}[t]{0.3\textwidth}
    \includegraphics[width=\textwidth]{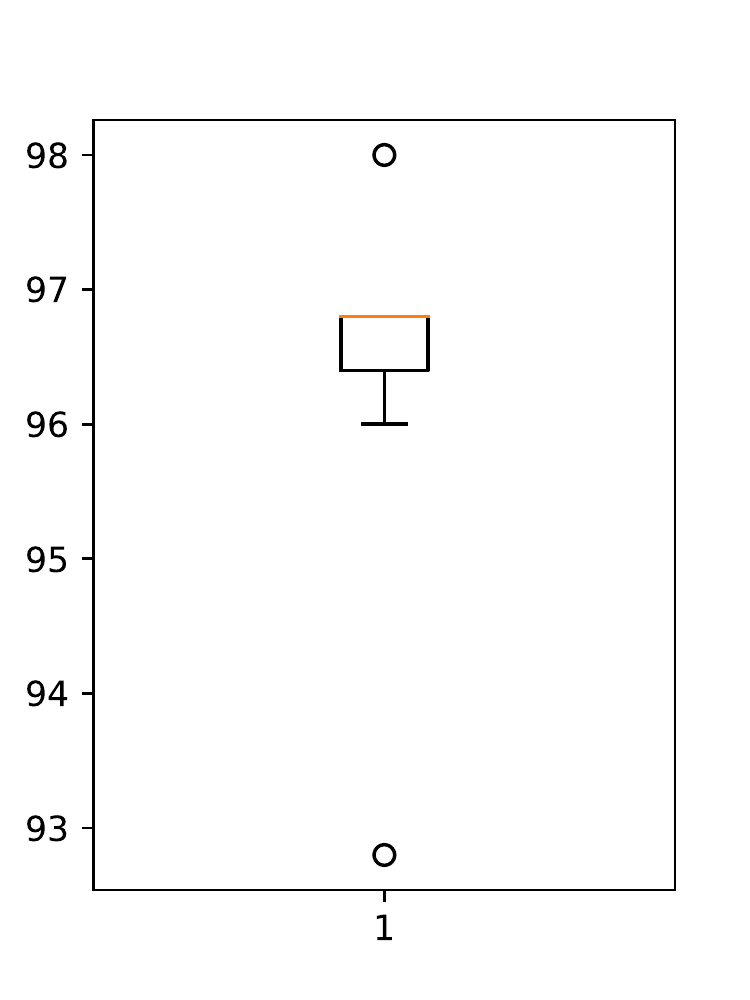}
    \caption{Manually created graphs}
\end{subfigure}\hspace{\fill} 
\begin{subfigure}[t]{0.3\textwidth}
    \includegraphics[width=\linewidth]{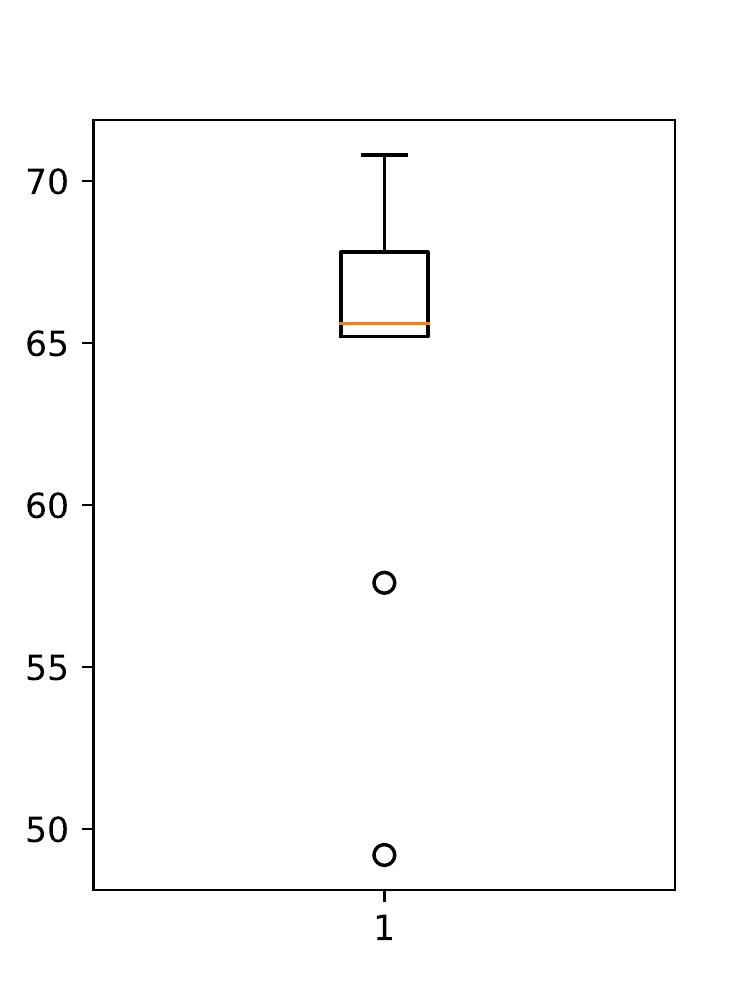}
    \caption{No graphs appended to original context}
\end{subfigure}\hspace{\fill}
\begin{subfigure}[t]{0.3\textwidth}
    \includegraphics[width=\linewidth]{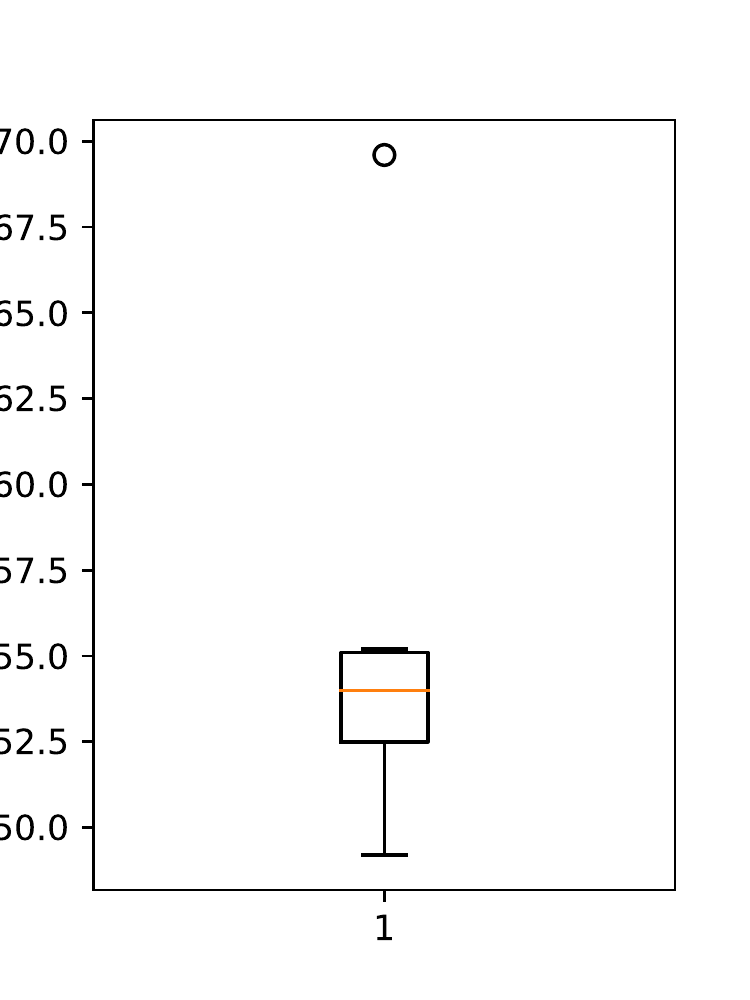}
    \caption{Approach 2}
\end{subfigure}\hspace{\fill} 
\begin{subfigure}[t]{0.30\textwidth}
    \includegraphics[width=\linewidth]{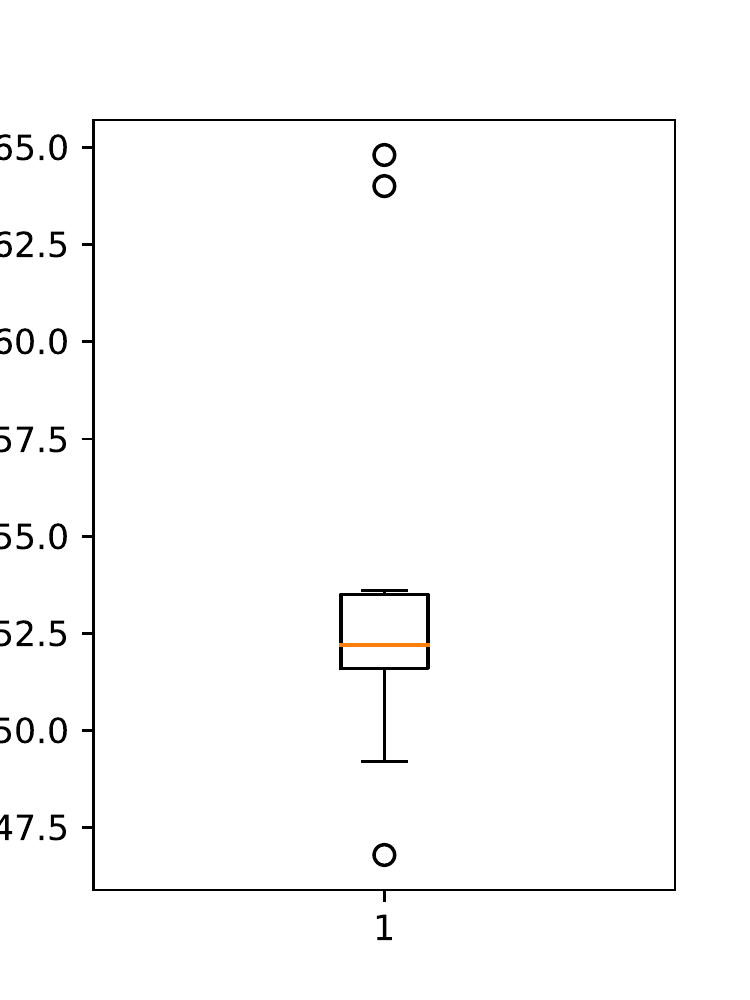}
    \caption{Approach 1}
\end{subfigure}\hspace{\fill}
\begin{subfigure}[t]{0.30\textwidth}
    \includegraphics[width=\linewidth]{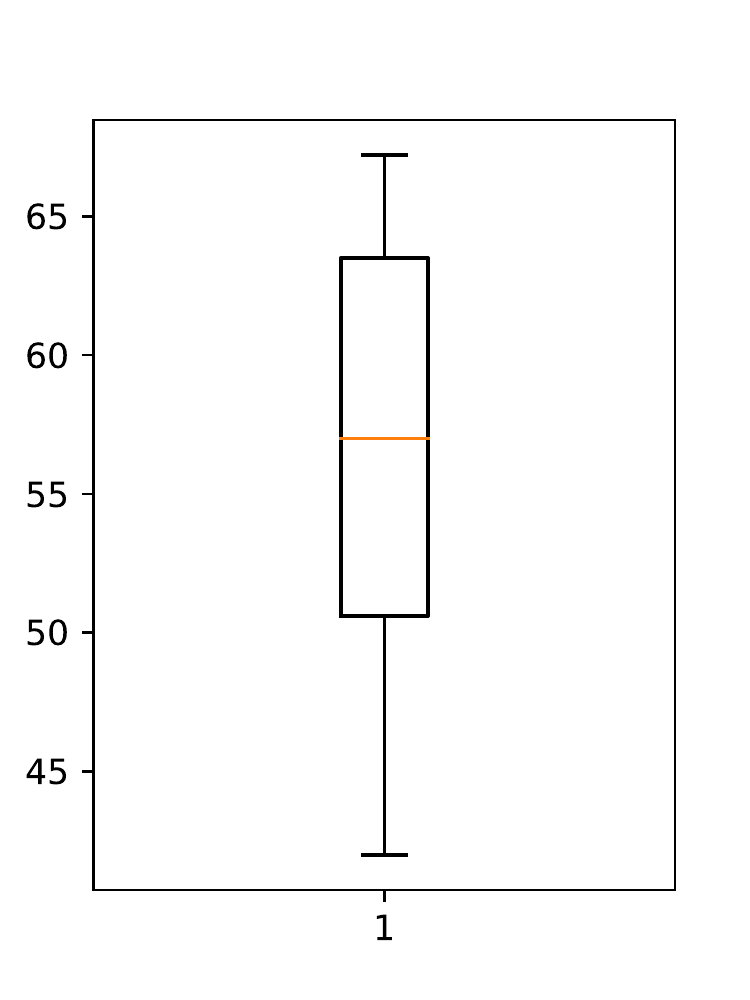}
    \caption{Approach 3}
\end{subfigure}\hspace{\fill} 
\begin{subfigure}[t]{0.30\textwidth}
    \includegraphics[width=\linewidth]{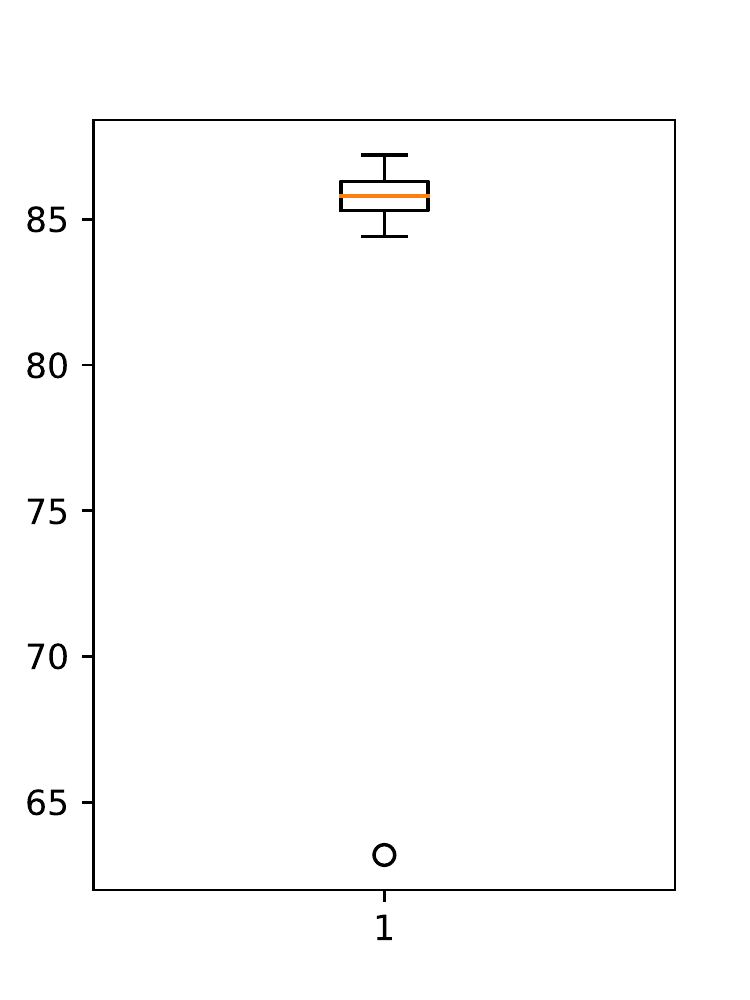}
    \caption{Approach 3-G}
\end{subfigure}

\caption{Outliers from the different runs for all graph configurations for COPA-SSE. Circular dots mark outliers that were removed, if any.}

\end{figure*}

\end{document}